\documentclass{article}
\usepackage{spconf,amsmath,graphicx}

\usepackage{array,booktabs,amsfonts}
\usepackage{algorithm}
\usepackage{algorithmic}
\usepackage{euscript,mathrsfs}

\title{Competitive learning for achieving content-specific filters in video
coding for machines }
\name{Honglei Zhang, Jukka I. Ahonen, Nam Le, Ruiying Yang, Francesco Cricri}
\address{Nokia Technologies, Finland}
\begin{document}
%
\maketitle
\begin{abstract}

This paper investigates the efficacy of jointly optimizing content-specific
post-processing filters to adapt a human-oriented video/image codec
into a codec suitable for machine vision tasks. By observing that
artifacts produced by video/image codecs are content-dependent, we
propose a novel training strategy based on competitive learning principles.
This strategy assigns training samples to filters dynamically, in
a fuzzy manner, which further optimizes the winning filter on the
given sample. Inspired by simulated annealing optimization techniques,
we employ a softmax function with a temperature variable as the weight
allocation function to mitigate the effects of random initialization.
Our evaluation, conducted on a system utilizing multiple post-processing
filters within a Versatile Video Coding (VVC) codec framework, demonstrates
the superiority of content-specific filters trained with our proposed
strategies, specifically, when images are processed in blocks. Using
VVC reference software VTM 12.0 as the anchor, experiments on the
OpenImages dataset show an improvement in the BD-rate reduction from
-41.3\% and -44.6\% to -42.3\% and -44.7\% for object detection and
instance segmentation tasks, respectively, compared to independently
trained filters. The statistics of the filter usage align with our
hypothesis and underscore the importance of jointly optimizing filters
for both content and reconstruction quality. Our findings pave the
way for further improving the performance of video/image codecs.

\end{abstract}

\section{Introduction}

\begin{figure}
\begin{centering}
\includegraphics[width=8cm]{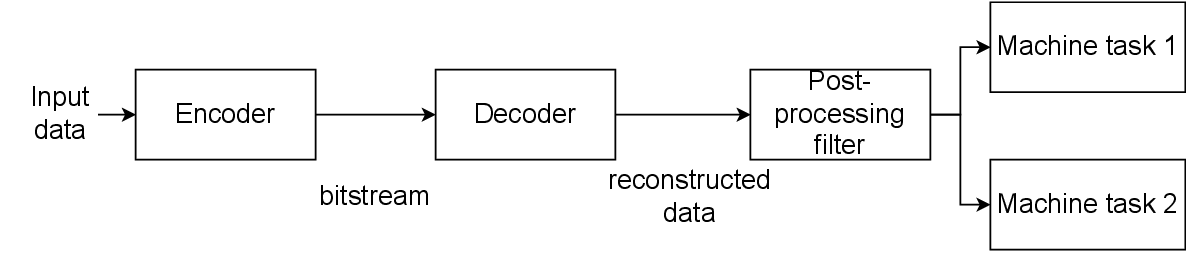}
\par\end{centering}
\caption{A VCM system with a conventional codec and multiple post-processing
filters \label{fig:vcm_system}}
\end{figure}

Video and image compression has been one of the most important technologies
for the industry with broad applications \cite{sullivan2005videocompression,bross2021overview}.
Recent advances in deep learning technologies have greatly boosted
the importance of video coding for machines (VCM), where the reconstructed
videos/images are consumed by machines to perform vision tasks such
as object detection, instance segmentation, and object tracking \cite{duan2020videocoding,yang2021videocoding,gao2021recentstandard}.
Among many other technologies proposed for VCM, previous research
has shown that post-processing filter can significantly improve the
performance of conventional codecs, which are optimized for human
beings, when used for machine tasks \cite{ahonen2021learned,ahonen2021imagecoding}.
Figure \ref{fig:vcm_system} shows a VCM codec with a conventional
video/image codec enhanced by multiple post-processing filters for
machine tasks. 

A video/image codec compresses an input video/image at various compression
ratios to generate reconstructed videos/images with different qualities
\cite{bross2020versatile,marpe2006theh264mpeg4}. Higher quality reconstructed
data is generated with lower compression ratios, i.e., higher bit
rate, and vice versa. In conventional video/image codecs, the compression
ratio is normally controlled by a quantization parameter (QP). It
is well known that different compression ratios may generate different
compression artifacts. For example, blurring and blocking artifacts
are more visible in low-quality reconstructions \cite{dong2015compression}.
To adapt different artifacts in the reconstructed data generated with
different compression ratios, a quality indicator variable, e.g.,
the QP value, may be given as an input to a post-processing filter
\cite{santamaria2021contentadaptive} as an indication of the type
of the artifacts. A post-processing filter is typically trained using
the reconstructed data of varying qualities along with the corresponding
quality indicator as inputs, providing additional information to enable
the filter to adapt to the input quality more effectively. In an alternative
approach, more than one post-processing filters are trained and each
post-processing filter is targeted to the reconstructed data of a
predefined quality range. At the inference stage, when processing
input data, the decoder selects the post-processing filter that has
been trained with the quality range encompassing the quality of the
sample \cite{zhang2020enhancing,li2017cnnbased}. This approach improves
the performance of the system without increasing the computational
complexity at the inference time compared to the one-filter approach. 

However, it is observed that the compression artifacts are not only
dependent on the compression ratio, e.g., QP values. More importantly,
the content of the input data also greatly impacts the artifacts.
For example, the size and strength of blocking artifacts may differ
significantly in homogeneous and heterogeneous regions due to the
difference in the coding unit sizes. In another example, ringing artifacts
may only appear in regions with regular patterns. For VCM, to improve
task performance, a post-processing filter may perform different operations
on different regions, depending on the content of the regions. For
example, a post-processing filter may smooth the background regions
such as sky and grass regions, to reduce false detection, and enhance
edges and patterns in object regions. Training multiple post-processing
filters to adapt the content of the input data requires dividing the
training data into different categories and assigning samples in each
category to a post-processing filter. However, defining the categories
would require prior knowledge, which is normally a difficult task.
To address these difficulties, we propose a joint training strategy
inspired by competitive learning to train multiple content-dependent
post-processing filters simultaneously. We evaluated the proposed
training strategy on a video codec for VCM with a conventional video
codec enhanced by post-processing filters. 

\section{Methods}

\subsection{Problems}

Post-processing filter is an efficient technology to improve the quality
of the reconstructed video for a conventional video/image codec, such
as VVC and HEVC \cite{santamaria2021contentadaptive,zhang2020enhancing,li2017cnnbased,yang2022lowprecision}.
A post-processing filter is normally trained by minimizing a loss
function, defined by 
\begin{equation}
L=\mathbb{E}_{x}D\left(F\left(\hat{x}|w\right),x\right),\label{eq:single_pp_train}
\end{equation}
where $x$ is uncompressed data, $\hat{x}$ is the reconstructed data
output by the conventional codec, $w$ is the learnable parameters
of the post-processing filter, $F(\cdot|w)$ is the output of the
post-processing filter, $D(\cdot)$ is a distortion measurement and
$\mathbb{E}_{x}(\cdot)$ is the expected value given random variable
$x$. The distortion measurement may be mean squared error (MSE),
structural similarity index measure (SSIM) \cite{nilsson2020understanding}
or alike for human consumption. In the case of video/image codec for
machines, proxy loss, e.g., MSE loss of the feature maps obtained
from a proxy network from uncompressed data and the output of the
post-processing filter, may be used as the distortion measurement
\cite{ahonen2021learned,le2021imagecoding}. A proxy network is normally
the backbone of a pretrained neural network model for a typical computer
vision task, such as instance segmentation. 

To obtain multiple post-processing filters with each filter dedicated
to a specific content, the parameters of the post-processing filters
can be derived by optimizing the loss function
\begin{equation}
L=\mathbb{E}_{x}\min_{j=1}^{M}\left(D\left(F\left(\hat{x}|w_{j}\right),x\right)\right),\label{eq:multi_pp_loss}
\end{equation}
 where $M$ is the number of post-processing filters, $w_{j}$ for
$j=1,\cdots,M$ is the learnable parameters of the $j$-th filter.
The $M$ post-processing filters can be derived by minimizing Eq.
\ref{eq:multi_pp_loss} with respect to $w_{1},w_{2},\cdots,w_{M}$.
However, minimizing Eq. \ref{eq:multi_pp_loss} directly is difficult
due to the nonlinear nature of the filters and the minimization function.
The training is highly sensitive to the initialization values of the
learnable parameters. 

Note that our target is to assign training samples to post-processing
filters such that similar samples are assigned to the same filter.
Stochastic gradient descent (SGD) is normally applied to train post-processing
filters, which are deep convolutional neural networks. Filters are
optimized at each iteration step, resulting in better performance,
for the samples in a training batch. By minimizing Eq. \ref{eq:multi_pp_loss},
the filter with the minimal distortion measurement for a sample gets
improved for that sample in one training iteration. As the parameters
of the multiple post-processing filters are randomly initialized,
the sample assignment to the filters is also random, thus resulting
in inefficient training. 

To alleviate the training difficulties, we replace the minimization
function in Eq. \ref{eq:multi_pp_loss} by a linear function to derive
a relaxation of the loss as
\begin{equation}
L=\mathbb{E}_{x}\left(\sum_{j=1}^{M}\alpha_{j}(\hat{x})D\left(F\left(\hat{x}|w_{j}\right),x\right)\right),\label{eq:min_sum_loss}
\end{equation}
subject to 
\begin{equation}
\sum_{j=1}^{M}\alpha_{j}(\hat{x})=1\;\mathrm{and}\ \alpha_{j}\left(\hat{x}\right)\ge0,\label{eq:min_sum_loss_subject_to}
\end{equation}
where $\alpha_{j}(\hat{x})$ is the weight for the $j$-th filter
for reconstructed data $\hat{x}$. When $\boldsymbol{\alpha}(\hat{x})=\left[\alpha_{1}(\hat{x}),\cdots,\alpha_{M}(\hat{x})\right]^{T}$
is a one-hot vector, i.e., one element in the vector has the value
of 1 and all other elements are zeros, minimizing Eq. \ref{eq:min_sum_loss}
is equivalent to minimizing the distortion loss of the filter with
the weight of one, i.e., assigning the sample $\hat{x}$ to the selected
filter. The optimization of Eq. \ref{eq:min_sum_loss} is an approximation
of assigning each sample to the filters during the optimization process
through a weighting strategy. 

\subsection{Competitive learning and annealing}

Following the principle of competitive learning \cite{fritzke1997somecompetitive,kohonen2013essentials},
we design a training strategy where the filters compete with each
other for the right to process a sample. The filter that obtains a
lower distortion measurement is assigned a higher weight in Eq. \ref{eq:min_sum_loss},
resulting in better performance for that sample after the optimization
step. Let $l_{j}=D\left(F\left(\hat{x}|w_{j}\right),x\right)$ be
the distortion measurement of filter $j$ for reconstructed data $\hat{x}$.
Inspired by the simulated annealing method \cite{aarts2005simulated},
we set 
\begin{equation}
\alpha_{j}\left(\hat{x}\right)=\frac{\exp\left(-\frac{l_{j}}{T}\right)}{\sum_{k=1}^{M}\exp\left(-\frac{l_{k}}{T}\right)},\label{eq:annealing_weight}
\end{equation}
where $T$ is the temperature variable that controls ``hardness''
of the assignment. Note that Eq. \ref{eq:annealing_weight} is a softmax
function with temperature $T$. When $T\rightarrow\infty$, $\alpha_{j}\rightarrow\frac{1}{M}$,
indicating the same weight for all filters. When $T\rightarrow0$,
$\boldsymbol{\alpha}\left(\hat{x}\right)$ is close to a one-hot vector,
indicating assigning all weights to the winning filter. Given a dataset
with $N$ training samples $\left(x_{i},\hat{x}_{i}\right)$ for $i=1,\cdots,N$,
the objective function for the dataset is 
\begin{equation}
L=\sum_{i=1}^{N}\sum_{j=1}^{M}\alpha_{j}(\hat{x}_{i})D\left(F\left(\hat{x}_{i}|w_{j}\right),x_{i}\right)\label{eq:annealing_loss}
\end{equation}

During the training, we gradually reduce temperature $T$ from the
initial temperature $T_{0}$. For simplicity, we choose a staircase
function as the cooling schedule for the temperature variable $T$.
The cooling function is defined as
\begin{equation}
T=T_{0}\beta^{-\left\lfloor \frac{k}{K}\right\rfloor },\label{eq:temperature_drop}
\end{equation}
 where $k$ is the epoch number, $K$ is the drop step, $\beta$ is
the base scale factor, and $\left\lfloor \cdot\right\rfloor $ is
floor function. According to the annealing function defined by Eq.
\ref{eq:temperature_drop}, the temperature $T$ drops $\beta$ times
every $K$ epochs. $T_{0}$, $K$ and $\beta$ are hyperparameters
of the training. 

\subsection{Filter architecture}

\begin{figure*}[tp]
\begin{centering}
\includegraphics[width=18cm]{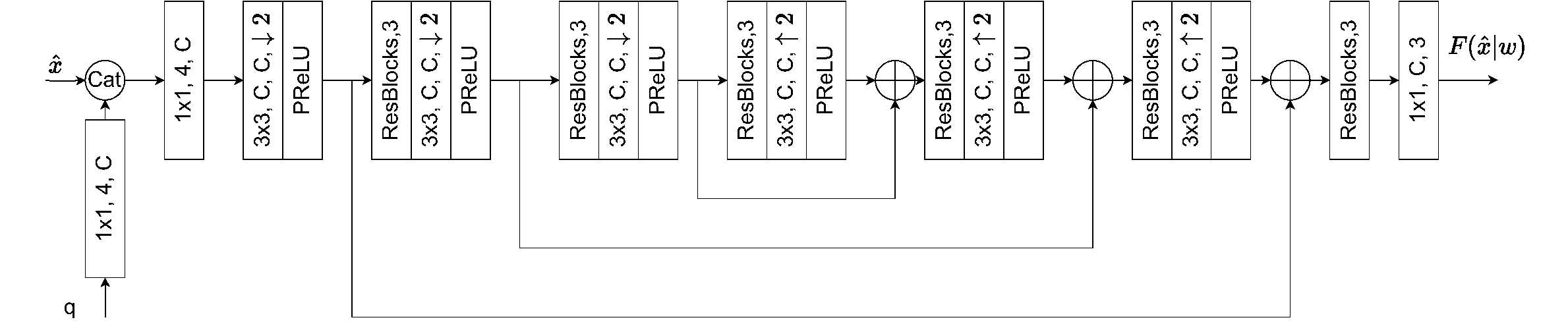}
\par\end{centering}
\caption{Neural network architecture of post-processing filters. \label{fig:filter_architecture}}
\end{figure*}

The architecture of the post-processing filter used in our experiments
incorporates an autoencoder backbone with lateral connections. The
detailed architecture is shown in Figure \ref{fig:filter_architecture},
where $\hat{x}$ is reconstructed video from a conventional video/image
codec, $F\left(\hat{x}|w\right)$ is the output of the filter given
learnable parameters $w$, $q$ is an indicator vector for the QP
value of the reconstructed data. Notation $k\times k,C_{1},C2$ indicates
a 2D convolution with kernel size $k$, input channel size $C_{1}$
and output channel size $C_{2}$. $\downarrow2$ indicates a convolution
operation with stride 2, which performs the down-sampling operation,
and $\uparrow2$ indicates a transposed convolution operation with
stride 2, which performs the up-sampling operation. ``ResBlocks,3''
represents a sequence of 3 residual blocks \cite{he2016identity},
and ``Cat'' in a circle represents a concatenation operation that
concatenates input tensors along channel dimension. 

\section{Experiments}

\subsection{Main results}

In our experiments, post-processing filters are applied to the reconstructed
data from a VVC codec \cite{bross2020versatile} to improve the performance
for machine tasks \cite{ahonen2021learned}. The system architecture
is shown in Figure \ref{fig:vcm_system}. 

To demonstrate the performance of the proposed method, named as \emph{joint-filters},
we compare the multiple post-processing filters trained using the
proposed training strategy with two baseline methods. With the first
baseline method, named as \emph{one-filter}, a post-processing filter
was trained on the data of various qualities. The filter has the same
network architecture as the one illustrated in Figure \ref{fig:filter_architecture}.
The QP value of reconstructed data is also given as an input to the
filter. With the second baseline method, named as \emph{independent-filters},
multiple post-processing filters were trained independently and each
post-processing filter was trained on a range of QP values. At the
inference time, each reconstructed data is assigned to the filter
trained with the range of the QP values that includes the QP value
of the reconstructed data. We evaluate the filters trained with the
proposed method and the competing methods on object detection and
instance segmentation tasks on the OpenImages dataset \cite{kuznetsova2020theopen}
following the evaluation procedure defined by the MPEG VCM standardization
working group \cite{iso/iecjtc1/sc29/wg04/n3112023commontest}. 

The video/image codec used in the experiments is VTM 12.0 and the
input images are compressed using 7 QP values: 22, 27, 32, 37, 42,
47, and 52. 30k training images were randomly selected from the training
split of the OpenImages V6 \cite{kuznetsova2020theopen}. Each of
the test sets for the object detection and instance segmentation tasks
contains 5k images. The machine tasks are evaluated using the mean
Average Precision scores (mAP). The compression performance of a codec
is measured using Bjontegaard delta bit rate (BD-rate) scores \cite{iso/iecjtc1/sc29/wg04/n3112023commontest},
which quantitatively measure the difference between the rate-distortion
(RD) curve of a test codec with the RD curve of an anchor codec. In
our experiments, the anchor codec is VTM 12.0 without a post-processing
filter. With the \emph{one-filter} baseline, the post-processing filter
was trained with all training data for 150 epochs. With the \emph{independent-filters}
baseline, 4 filters were trained independently. The filters are separated
with QP values in the range $[22,27]$, $[32,37]$, $[42,47]$, and
$[52]$ respectively. Each filter was trained for 100 epochs. With
the proposed training strategy, 4 filters were trained jointly for
100 epochs. In all experiments, the filters were trained using image
patches of size $256\times256$ and Adam optimizer \cite{kingma2014adama}
with a learning rate 2E-4. Note that in these experiments, the bit
rates of the data points from the test methods are the same as the
anchor since the test codec used the same encoder as the anchor codec. 

Considering that the content of a natural image is normally not homogeneous,
for example, an image may contain various regions with different smoothness
and patterns. Applying content-specific filters to the whole picture
does not exploit the advantages of jointly trained filters. We further
studied block-wise processing, partitioning an input reconstructed
image into blocks with the size of $256\times256$ or $128\times128$
pixels and applying the most suitable filter to each block. To determine
the filter for a block, the encoder calculates the proxy loss of all
candidate filters and selects the one with the lowest loss value.
The encoder signals the index of the selected filters to the decoder
along with the bitstream, resulting in 2 bits per block overhead.
At the decoder side, each block is processed by the selected filter
indicated by the filter index signaled by the encoder. We collected
the results from both jointly trained filters and independently trained
filters when the reconstructed data is processed in a block-wise manner. 

Table \ref{tab:main_results} shows the BD-rate of the object detection
and instance segmentation tasks of the proposed method and the baseline
methods against VTM 12.0 anchor. The bold fonts show the best BD-rate
values. 

\begin{table*}
\caption{Comparison of the baseline methods and the proposed training strategy
against VTM 12.0 anchor.\label{tab:main_results}}

\smallskip{}

\centering{}%
\begin{tabular}{|c|c|c|}
\hline 
method & Detection BD-rate  & Segmentation BD-rate\tabularnewline
\hline 
\hline 
one-filter & -38.3\% & -39.4\%\tabularnewline
\hline 
4 independent-filters & -39.82\% & -42.75\%\tabularnewline
\hline 
4 joint-filters & -39.1\% & -40.1\%\tabularnewline
\hline 
4 independent-filter (block-wise, $256\times256$) & -40.7\% & -43.7\%\tabularnewline
\hline 
4 joint-filters (block-wise, $256\times256$) & -42.2\% & -43.6\%\tabularnewline
\hline 
4 independent-filters (block-wise, $128\times128$) & -41.3\% & -44.6\%\tabularnewline
\hline 
\textbf{4 joint-filters (block-wise, }$128\times128$\textbf{)} & \textbf{-42.3\%} & \textbf{-44.7\%}\tabularnewline
\hline 
\end{tabular}
\end{table*}

The BD-rates for object detection and instance segmentation tasks
in Table \ref{tab:main_results} show that 4 filters trained independently
or jointly outperform one filter trained for all QP values. It should
be noted that the computational complexity of 4 filters is the same
as the complexity of one filter at the inference time, although the
number of parameters is 4 times that of the single-filter system.
The results also show that independently trained filters are better
than jointly trained filters when the reconstructed image is processed
as a whole. This demonstrates that each filter in jointly trained
filters is specialized for certain content, thus, processing the whole
image using one filter does not guarantee a good performance. The
independently trained filters are quality-specific, which works better
in this scenario. When the input images are processed with 256x256
blocks, jointly trained filters are better than the independently
trained filters on the object detection task, and slightly worse on
segmentation task. When the images are processed in 128x128 blocks,
jointly trained filters prove superior to independently trained filters
on all machine tasks. 

\begin{figure}
\begin{centering}
\includegraphics[width=8cm]{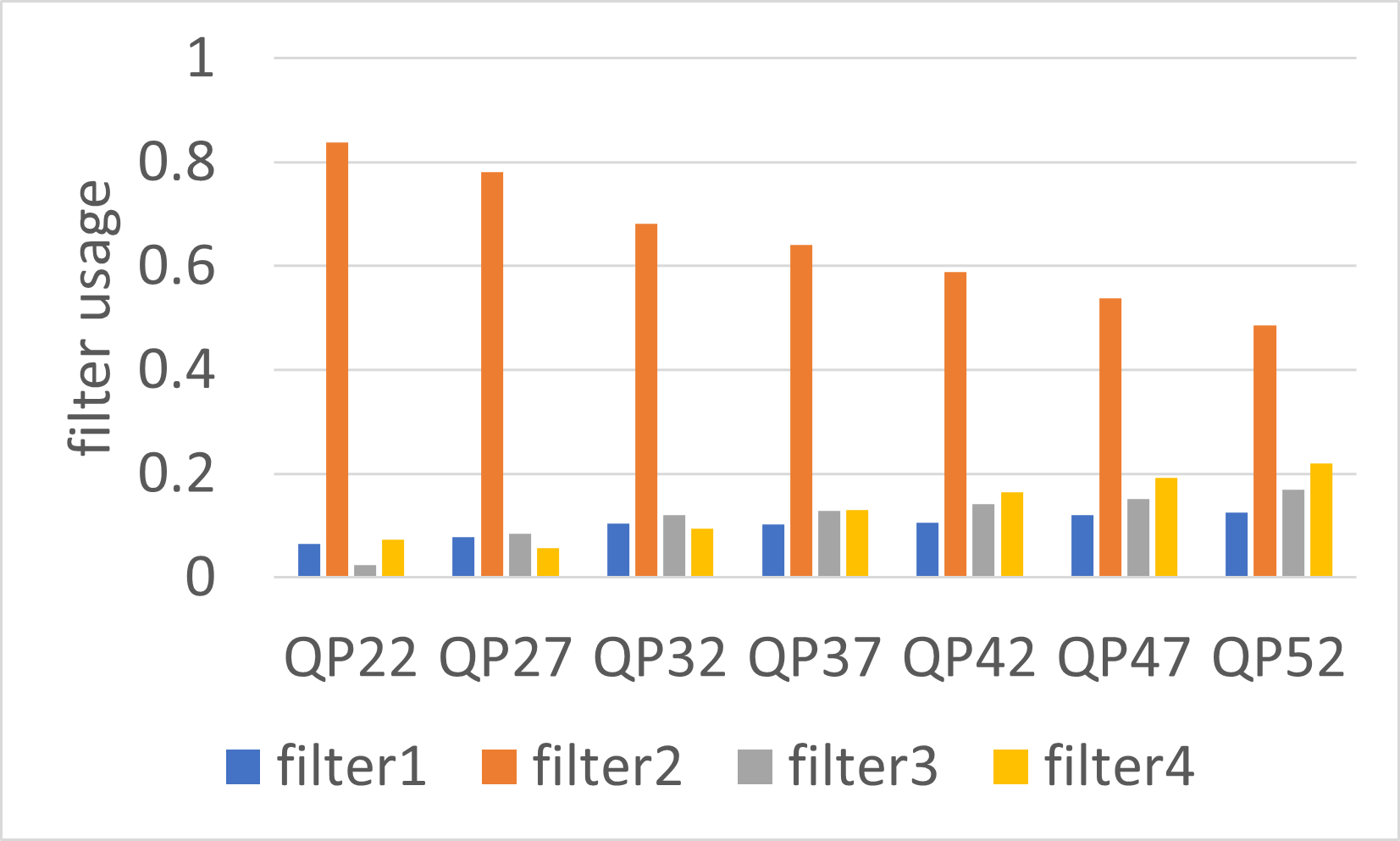}
\par\end{centering}
\caption{Filter usage statistics against QP values. \label{fig:filter_usage}}
\end{figure}

We also collected the statistics of the usage of each individual filter
in jointly trained filters to demonstrate sample assignment, e.g.,
blocks in reconstructed images, with regard to the quality of the
reconstructed images. The usage percentage of each filter in jointly
trained filters against the sample QP values is shown in Figure \ref{fig:filter_usage}

The results indicate that filter2 is dominant among all samples, as
it processes more than 60\% of the blocks across all QP values. As
demonstrated in the figure, with increasing QP values, the usage of
filter2 decreases from over 80\% to about 50\%, while the usage of
other filters increases. These findings indicate that the jointly
trained filters are optimized for the content of the samples rather
than being solely based on the QP values.

\subsection{Study of training hyperparameters}

The proposed method uses a staircase function as the cooling schedule
and drops the temperature $T$ gradually during the training. We empirically
determined the parameter $K$ defined in Eq. \ref{eq:temperature_drop},
which controls the speed of the cooling. To determine the optimal
parameter $K$, we tested 4 different values of $K$: $K=1,$ $K=3,$
$K=5$, and $K=10$. The filters were trained for 20 epochs and their
performance was evaluated on the test data. In all experiments, we
set $\beta=10$ and $T_{0}=1$. The results are shown in Table \ref{tab:effect_of_K}

\begin{table}[h]
\caption{Performance study of parameter $K$.\label{tab:effect_of_K}}

\smallskip{}

\begin{centering}
\begin{tabular}{|c|c|c|}
\hline 
$K$ & Detection BD-rate & Segmentation BD-rate\tabularnewline
\hline 
\hline 
1 epoch  & -33.13\% & -36.54\%\tabularnewline
\hline 
3 epochs  & -33.81\% & -35.680\%\tabularnewline
\hline 
5 epochs  & \textbf{-35.212\%} & \textbf{-39.532\%}\tabularnewline
\hline 
10 epochs & -33.426\% & -36.960\%\tabularnewline
\hline 
\end{tabular}
\par\end{centering}
\vspace{-2mm}
\end{table}

Table \ref{tab:effect_of_K} indicates that the system achieves the
best performance when $K=5$. When $K$ is too large, the temperature
decreases slowly, resulting in the filters being trained on all input
samples, therefore limiting the system performance due to the capacity
of each individual filter. Conversely, when $K$ is too small, the
temperature decreases too fast, causing randomly assigning the training
samples to each individual filter, thus, hindering the performance
of the system. This observation also showcases the inferior performance
when Eq. \ref{eq:multi_pp_loss} is optimized directly. 

\section{Conclusion}

Post-processing filter is an efficient and convenient technique for
enhancing the performance of a video/image codec or adapting an existing
video/image codec for other purposes, such as converting a conventional
video/image codec designed for human into one suitable for machine
vision tasks. Existing techniques utilize multiple post-processing
filters within a system and optimize each filter for a range of reconstruction
qualities, for example, a range of QP values. We observe that the
artifacts caused by a video/image codec are also content-specific.
A system with multiple post-processing filters could perform better
when the filters are jointly optimized. In this paper, we proposed
a training strategy that optimizes post-processing filters jointly
following the principle of competitive learning. During the training
stage, a training sample is assigned, in a fuzzy manner, to the filters
based on the performance of each filter on the sample, such that the
winning filter is further optimized for the sample to achieve a better
performance. To mitigate the randomness caused by random initialization,
inspired by the simulated annealing technique, we assign samples to
each filter using a soft-max function with temperature as the weight
allocation function and gradually decrease the temperature during
the training. We evaluated the proposed training strategy with a system
using multiple post-processing filters on a VVC codec \cite{bross2020versatile,ahonen2021learned}.
Following the evaluation procedure defined by the MPEG VCM standardization
working group \cite{iso/iecjtc1/sc29/wg04/n3112023commontest}, we
assessed the proposed system on the OpenImages dataset with the object
detection and instance segmentation machine tasks. Experiments show
that the content-specific filters trained with the proposed strategies
outperform the filters trained independently when images are processed
in blocks. The usage statistics of the jointly trained filters confirm
that the filters are specific to content and reconstruction quality
rather than solely focusing on reconstruction quality. 

In this paper, we demonstrated the effectiveness of jointly trained
post-processing filters for VCM codecs. Further experiments will be
performed to validate the efficiency of the proposed training strategy
for conventional video codecs.

\bibliographystyle{IEEEbib}
\bibliography{clean}

\begin{thebibliography}{10}

\bibitem{sullivan2005videocompression}
G.J. Sullivan and T.~Wiegand,
\newblock ``Video {Compression} - {From} {Concepts} to the {H}.264/{AVC}
  {Standard},''
\newblock {\em Proceedings of the IEEE}, vol. 93, no. 1, pp. 18--31, Jan. 2005.

\bibitem{bross2021overview}
Benjamin Bross, Ye-Kui Wang, Yan Ye, Shan Liu, Jianle Chen, Gary~J. Sullivan,
  and Jens-Rainer Ohm,
\newblock ``Overview of the {Versatile} {Video} {Coding} ({VVC}) {Standard} and
  its {Applications},''
\newblock {\em IEEE Transactions on Circuits and Systems for Video Technology},
  vol. 31, no. 10, pp. 3736--3764, Oct. 2021.

\bibitem{duan2020videocoding}
Lingyu Duan, Jiaying Liu, Wenhan Yang, Tiejun Huang, and Wen Gao,
\newblock ``Video {Coding} for {Machines}: {A} {Paradigm} of {Collaborative}
  {Compression} and {Intelligent} {Analytics},''
\newblock {\em IEEE Transactions on Image Processing}, vol. 29, pp. 8680--8695,
  2020.

\bibitem{yang2021videocoding}
Wenhan Yang, Haofeng Huang, Yueyu Hu, Ling-Yu Duan, and Jiaying Liu,
\newblock ``Video {Coding} for {Machine}: {Compact} {Visual} {Representation}
  {Compression} for {Intelligent} {Collaborative} {Analytics},'' Oct. 2021.

\bibitem{gao2021recentstandard}
Wen Gao, Shan Liu, Xiaozhong Xu, Manouchehr Rafie, Yuan Zhang, and Igor Curcio,
\newblock ``Recent {Standard} {Development} {Activities} on {Video} {Coding}
  for {Machines},'' May 2021.

\bibitem{ahonen2021learned}
Jukka~I. Ahonen, Ramin~G. Youvalari, Nam Le, Honglei Zhang, Francesco Cricri,
  Hamed~Rezazadegan Tavakoli, Miska~M. Hannuksela, and Esa Rahtu,
\newblock ``Learned {Enhancement} {Filters} for {Image} {Coding} for
  {Machines},''
\newblock in {\em 2021 {IEEE} {International} {Symposium} on {Multimedia}
  ({ISM})}, Nov. 2021, pp. 235--239.

\bibitem{ahonen2021imagecoding}
Jukka Ahonen,
\newblock ``Image coding for machines: {Deep} learning based post-processing
  filters,''
\newblock M.S. thesis, 2021.

\bibitem{bross2020versatile}
Benjamin Bross, Jianle Chen, Shan Liu, and Ye-Kui Wang,
\newblock ``Versatile {Video} {Coding} ({Draft} 8),''
\newblock {\em Joint Video Experts Team (JVET), Document JVET-Q2001}, Jan.
  2020.

\bibitem{marpe2006theh264mpeg4}
D.~Marpe, T.~Wiegand, and G.J. Sullivan,
\newblock ``The {H}.264/{MPEG4} advanced video coding standard and its
  applications,''
\newblock {\em IEEE Communications Magazine}, vol. 44, no. 8, pp. 134--143,
  Aug. 2006.

\bibitem{dong2015compression}
Chao Dong, Yubin Deng, Chen~Change Loy, and Xiaoou Tang,
\newblock ``Compression artifacts reduction by a deep convolutional network,''
\newblock in {\em Proceedings of the {IEEE} international conference on
  computer vision}, 2015, pp. 576--584.

\bibitem{santamaria2021contentadaptive}
Maria Santamaria, Yat-Hong Lam, Francesco Cricri, Jani Lainema, Ramin~G.
  Youvalari, Honglei Zhang, Miska~M. Hannuksela, Esa Rahtu, and Moncef Gaubbuj,
\newblock ``Content-adaptive convolutional neural network post-processing
  filter,''
\newblock in {\em 2021 {IEEE} {International} {Symposium} on {Multimedia}
  ({ISM})}, Nov. 2021, pp. 99--106.

\bibitem{zhang2020enhancing}
Fan Zhang, Chen Feng, and David~R. Bull,
\newblock ``Enhancing {VVC} {Through} {Cnn}-{Based} {Post}-{Processing},''
\newblock in {\em 2020 {IEEE} {International} {Conference} on {Multimedia} and
  {Expo} ({ICME})}, July 2020, pp. 1--6.

\bibitem{li2017cnnbased}
Chen Li, Li~Song, Rong Xie, and Wenjun Zhang,
\newblock ``{CNN} based post-processing to improve {HEVC},''
\newblock in {\em 2017 {IEEE} {International} {Conference} on {Image}
  {Processing} ({ICIP})}, Sept. 2017, pp. 4577--4580.

\bibitem{yang2022lowprecision}
Ruiying Yang, Maria Santamaria, Francesco Cricri, Honglei Zhang, Jani Lainema,
  Ramin~G. Youvalari, and Miska~M. Hannuksela,
\newblock ``Low-precision post-filtering in video coding,''
\newblock in {\em 2022 {IEEE} {International} {Symposium} on {Multimedia}
  ({ISM})}, Dec. 2022, pp. 137--140.

\bibitem{nilsson2020understanding}
Jim Nilsson and Tomas Akenine-M{\"o}ller,
\newblock ``Understanding {SSIM},''
\newblock {\em arXiv:2006.13846 [cs, eess]}, June 2020.

\bibitem{le2021imagecoding}
Nam Le, Honglei Zhang, Francesco Cricri, Ramin Ghaznavi-Youvalari, and Esa
  Rahtu,
\newblock ``Image {Coding} {For} {Machines}: an {End}-{To}-{End} {Learned}
  {Approach},''
\newblock in {\em {ICASSP} 2021 - 2021 {IEEE} {International} {Conference} on
  {Acoustics}, {Speech} and {Signal} {Processing} ({ICASSP})}, June 2021, pp.
  1590--1594.

\bibitem{fritzke1997somecompetitive}
Bernd Fritzke,
\newblock ``Some competitive learning methods,''
\newblock {\em Artificial Intelligence Institute, Dresden University of
  Technology}, vol. 100, 1997.

\bibitem{kohonen2013essentials}
Teuvo Kohonen,
\newblock ``Essentials of the self-organizing map,''
\newblock {\em Neural Networks}, vol. 37, pp. 52--65, Jan. 2013.

\bibitem{aarts2005simulated}
Emile Aarts, Jan Korst, and Wil Michiels,
\newblock ``Simulated {Annealing},''
\newblock in {\em Search {Methodologies}: {Introductory} {Tutorials} in
  {Optimization} and {Decision} {Support} {Techniques}}, Edmund~K. Burke and
  Graham Kendall, Eds., pp. 187--210. Springer US, Boston, MA, 2005.

\bibitem{he2016identity}
Kaiming He, Xiangyu Zhang, Shaoqing Ren, and Jian Sun,
\newblock ``Identity {Mappings} in {Deep} {Residual} {Networks},''
\newblock {\em arXiv:1603.05027 [cs]}, Mar. 2016.

\bibitem{kuznetsova2020theopen}
Alina Kuznetsova, Hassan Rom, Neil Alldrin, Jasper Uijlings, Ivan Krasin, Jordi
  Pont-Tuset, Shahab Kamali, Stefan Popov, Matteo Malloci, Alexander
  Kolesnikov, Tom Duerig, and Vittorio Ferrari,
\newblock ``The {Open} {Images} {Dataset} {V4}: {Unified} image classification,
  object detection, and visual relationship detection at scale,''
\newblock {\em IJCV}, 2020.

\bibitem{iso/iecjtc1/sc29/wg04/n3112023commontest}
ISO/IEC JTC 1/SC 29/WG~04/N 311,
\newblock ``Common test conditions for video coding for machines,'' Feb. 2023.

\bibitem{kingma2014adama}
Diederik~P. Kingma and Jimmy Ba,
\newblock ``Adam: {A} {Method} for {Stochastic} {Optimization},''
\newblock Dec. 2014.

\end{thebibliography}

\end{document}